%% file: main.tex
\documentclass[11pt,a4paper]{article}
\usepackage[hyperref]{ACL2020}
\usepackage{times}
\usepackage{latexsym}

\usepackage{microtype}
\usepackage[algo2e,ruled,vlined]{algorithm2e} 
\usepackage{setspace}
\usepackage{arydshln}
\usepackage{verbatim}
\usepackage{xcolor}  
\usepackage{soul}  
\input{settings.tex}

\input{math_commands.tex}

\definecolor{mygreen}{RGB}{28,124,53} 
\definecolor{mygreen1}{RGB}{202,234,202} 
\definecolor{myorange}{RGB}{155,127,14}
\definecolor{myorange1}{RGB}{249,206,175} 
\definecolor{myblue}{RGB}{211,221,237} 
\definecolor{myblue1}{RGB}{189,208,229} 
\definecolor{myred}{RGB}{242,192,194} 

\newcommand{\dashrule}[1][black]{%
  \color{#1}\rule[\dimexpr.2ex-.2pt]{4pt}{.4pt}\xleaders\hbox{\rule{2pt}{0pt}\rule[\dimexpr.2ex-.2pt]{4pt}{.4pt}}\hfill\kern0pt%
}
\aclfinalcopy 

\title{ERNIE-Gram: Pre-Training with Explicitly N-Gram Masked Language Modeling for Natural Language Understanding}

\author{
Dongling Xiao,~~Yukun Li,~~Han Zhang,~~Yu Sun,~~Hao Tian, \\
\textbf{Hua Wu} and \textbf{Haifeng Wang} \\
Baidu Inc., China\\
{\fontsize{12pt}{0}\selectfont\texttt{\{xiaodongling,liyukun01,zhanghan17,sunyu02,}} \\{\fontsize{12pt}{0}\selectfont \texttt{tianhao,wu\_hua,wanghaifeng\}@baidu.com}}
}

\date{}

\begin{document}
\maketitle
\begin{abstract}
Coarse-grained linguistic information, such as named entities or phrases, facilitates adequate-ly representation learning in pre-training. Previous works mainly focus on extending the objective of BERT's Masked Language Modeling (MLM) from masking individual tokens to contiguous sequences of $n$ tokens. We argue that such contiguously masking method neglects to model the intra-dependencies and inter-relation of coarse-grained linguistic information. As an alternative, we propose ERNIE-Gram, an explicitly $n$-gram masking method to enhance the integration of coarse-grained information into pre-training. In ERNIE-Gram, $n$-grams are masked and predicted directly using explicit $n$-gram identities rather than contiguous sequences of $n$ tokens. Furthermore, ERNIE-Gram employs a generator model to sample plausible $n$-gram identities as optional n-gram masks and predict them in both coarse-grained and fine-grained manners to enable comprehensive $n$-gram prediction and relation modeling. We pre-train ERNIE-Gram on English and Chinese text corpora and fine-tune on 19 downstream tasks. Experimental results show that ERNIE-Gram outperforms previous pre-training models like XLNet and RoBERTa by a large margin, and achieves comparable results with state-of-the-art methods. The source codes and pre-trained models have been released at \url{https://github.com/PaddlePaddle/ERNIE}.
\end{abstract}

\section{Introduction}
\label{sec:intro}

Pre-trained on large-scaled text corpora and fine-tuned on downstream tasks, self-supervised representation models \cite{gpt,bert,roberta,xlnet,albert,electra} have achieved remarkable improvements in natural language understanding (NLU). 
As one of the most prominent pre-trained models, BERT~\cite{bert} employs masked language modeling (MLM) to learn representations by masking individual tokens and predicting them based on their bidirectional context. However, BERT's MLM focuses on the representations of fine-grained text units (e.g. words or subwords in English and characters in Chinese), rarely considering the coarse-grained linguistic information (e.g. named entities or phrases in English and words in Chinese) thus incurring inadequate representation learning. 

Many efforts have been devoted to integrate coarse-grained semantic information by independently masking and predicting contiguous sequences of $n$ tokens, namely $n$-grams, such as named entities, phrases~\cite{ernie1}, whole words~\cite{bert-wwm} and text spans~\cite{spanbert}.
We argue that such contiguously masking strategies are less effective and reliable since the prediction of tokens in masked n-grams are independent of each other, which neglects the intra-dependencies of n-grams. Specifically, given a masked $n$-gram $\bm{w}\!=\!\{x_1,...,x_n\}, x\!\in\!\mathcal{V}_F$, we maximize $p(\bm{w})\!=\!\prod_{i=1}^np(x_i|\bm{c})$ for $n$-gram learning, where models learn to recover $\bm{w}$ in a huge and sparse prediction space $\mathcal{F}\!\in\!\mathbb{R}^{|\mathcal{V}_F|^n}$. Note that $\mathcal{V}_F$ is the fine-grained vocabulary\footnote{$\mathcal{V}_F$ contains $30$K BPE codes in BERT~\cite{bert} and $50$K subword units in RoBERTa~\cite{roberta}.} and $\bm{c}$ is the context. 

\begin{figure*}[t]
\setlength{\belowcaptionskip}{-0.5cm}
\setlength{\abovecaptionskip}{4pt}
\begin{center} 
\includegraphics[width=0.99\linewidth]{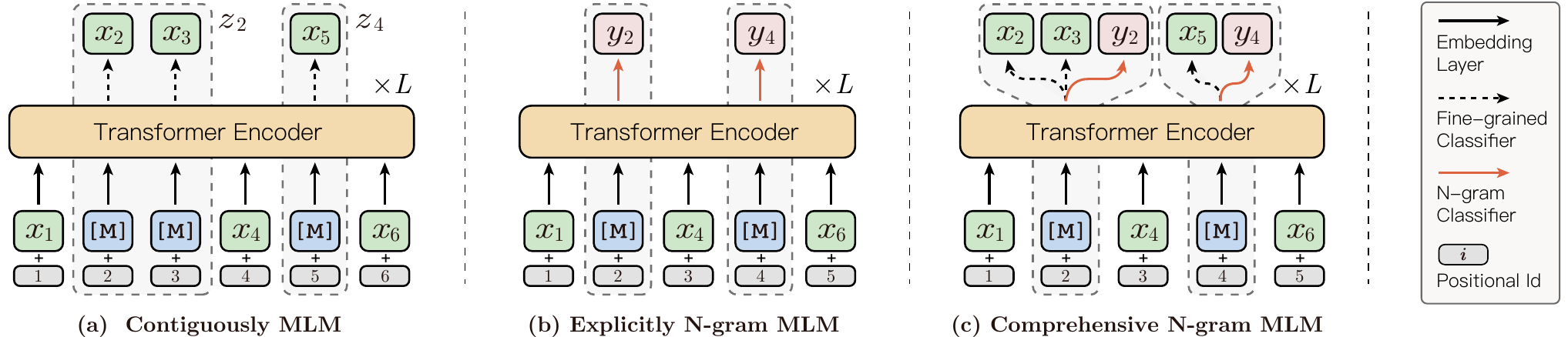}
\caption{Illustrations of different MLM objectives, where $x_i$ and $y_i$ represent the identities of fine-grained tokens and explicit $n$-grams respectively. Note that the weights of fine-grained classifier ($W_F\!\in\!\mathbb{R}^{h\times|\mathcal{V}_F|}$) and N-gram classifier ($W_N\!\in\!\mathbb{R}^{h\times|\langle\mathcal{V}_F, \mathcal{V}_N\rangle|}$) are not used in fine-tuning stage, where $h$ is the hidden size and $L$ is the layers.}
\label{overview}
\end{center} 
\end{figure*}

We propose ERNIE-Gram, an \textbf{explicitly $\bm{n}$-gram masked} language modeling method in which $n$-grams are masked with single {\tt[MASK]} symbols, and predicted directly using explicit $n$-gram identities rather than sequences of tokens, as depicted in Figure~\ref{overview}(b). The models learn to predict $n$-gram $\bm{w}$ in a small and dense prediction space $\mathcal{N}\!\!\in\!\mathbb{R}^{|\mathcal{V}_{N}|}$, where $\mathcal{V}_{N}$ indicates a prior $n$-gram lexicon\footnote{$\mathcal{V}_{N}$ contains $300$K $n$-grams, where $n\!\in\![2,4)$ in this paper, $n$-grams are extracted in word-level before tokenization.} and normally $|\mathcal{V}_{N}|\!\ll\!|\mathcal{V}_F|^n$. 
To learn the semantic of $n$-grams more adequately, we adopt a \textbf{comprehensive $\bm{n}$-gram prediction} mechanism, simultaneously predicting masked $n$-grams in coarse-grained (explicit $n$-gram identities) and fine-grained (contained token identities) manners with well-designed attention mask metrics, as shown in Figure~\ref{overview}(c). 

In addition, to model the semantic relationships between $n$-grams directly, we introduce an \textbf{enhanced $\bm{n}$-gram relation modeling} mechanism, masking $n$-grams with plausible $n$-grams identities sampled from a generator model, and then recovering them to the original $n$-grams with the pair relation between plausible and original $n$-grams. Inspired by ELECTRA~\cite{electra}, we incorporate the replaced token detection objective to distinguish original $n$-grams from plausible ones, which enhances the interactions 
between explicit $n$-grams and fine-grained contextual tokens.

In this paper, we pre-train ERNIE-Gram on both base-scale and large-scale text corpora (16GB and 160GB respectively) under comparable pre-training setting. Then we fine-tune ERNIE-Gram on 13 English NLU tasks and 6 Chinese NLU tasks. Experimental results show that ERNIE-Gram consistently outperforms previous well-performed pre-training models on various benchmarks by a large margin.

\section{Related Work\vspace{-4pt}}
\subsection{Self-Supervised Pre-Training for NLU\vspace{-3pt}} 
Self-supervised pre-training has been used to learn contextualized sentence representations though various training objectives. GPT~\cite{gpt} employs unidirectional language modeling (LM) to exploit large-scale corpora. BERT~\cite{bert} proposes masked language modeling (MLM) to learn bidirectional representations efficiently, which is a representative objective for pre-training and has numerous extensions such as RoBERTa~\cite{roberta}, \textsc{UniLM}~\cite{unilmv1} and ALBERT~\cite{albert}.
XLNet~\cite{xlnet} adopts permutation language modeling (PLM) to model the dependencies among predicted tokens. 
ELECTRA introduces replaced token detection (RTD) objective to learn all tokens for more compute-efficient pre-training. 

\subsection{Coarse-grained Linguistic Information Incorporating for Pre-Training\vspace{-4pt}} 
Coarse-grained linguistic information is indispensable for adequate representation learning.
There are lots of studies that implicitly integrate coarse-grained information by extending BERT's MLM to contiguously masking and predicting contiguous sequences of  tokens. For example, ERNIE~\cite{ernie1} masks named entities and phrases to enhance contextual representations, BERT-wwm~\cite{bert-wwm} masks whole Chinese words to achieve better Chinese representations, SpanBERT~\cite{spanbert} masks contiguous spans to improve the performance on span selection tasks. 

A few studies attempt to inject the coarse-grained $n$-gram representations into fine-grained contextualized representations explicitly, such as \textsc{Zen}~\cite{zen} and AMBERT~\cite{ambert}, in which additional transformer encoders and computations for explicit $n$-gram representations are incorporated into both pre-training and fine-tuning. \citealp{word-segmentation} demonstrate that explicit $n$-gram representations are not sufficiently reliable for NLP tasks because of $n$-gram data sparsity and the ubiquity of out-of-vocabulary $n$-grams. Differently, we only incorporate $n$-gram information by leveraging auxiliary $n$-gram classifier and embedding weights in pre-training, which will be completely removed during fine-tuning, so our method maintains the same parameters and computations as BERT.

\section{Proposed Method\vspace{-4pt}}
In this section, we present the detailed implementation of ERNIE-Gram, including $n$-gram lexicon $\mathcal{V}_{N}$ extraction in Section~\ref{sec3_2}, explicitly $n$-gram MLM pre-training objective in Section~\ref{sec3_3}, comprehensive $n$-gram prediction and relation modeling mechanisms in Section~\ref{sec3_4} and~\ref{sec3_5}.
\subsection{Background}
To inject $n$-gram information into pre-training, many works~\cite{ernie1,bert-wwm,spanbert} extend BERT's masked language modeling (MLM) from masking individual tokens to contiguous sequences of $n$ tokens. 
\paragraph{Contiguously MLM.} Given input sequence $\bm{x}\!=\!\{x_1,...,x_{|\bm{x}|}\}, x\!\in\!\mathcal{V}_F$ and $n$-gram starting boundaries $\bm{b}\!=\!\{b_1,...,b_{|\bm{b}|}\}$, let $\bm{z}=\{z_1,...,z_{|\bm{b}|-1}\}$ to be the sequence of $n$-grams, where $z_i\!=\!\bm{x}_{[b_i:b_{i+1})}$, 
MLM samples $15\%$ of starting boundaries from $\bm{b}$ to mask $n$-grams, donating ${\mathcal{M}}$ as the indexes of sampled starting boundaries, $\bm{z}_{{\mathcal{M}}}$ as the contiguously masked tokens, $\bm{z}_{{\setminus\mathcal{M}}}$ as the sequence after masking. As shown in Figure~\ref{overview}(a), $\bm{b}\!=\!\{1,2,4,5,6,7\},\bm{z}\!=\!\{x_1,\bm{x}_{[2:4)},x_4,x_5,x_6\},\mathcal{M}\!=\!\{2,4\}, \bm{z}_{\mathcal{M}}\!=\!\{\bm{x}_{[2:4)},x_5\},$ and $
\bm{z}_{{\setminus\mathcal{M}}}\!=\!\{x_1,\texttt{[M]},$ $\texttt{[M]},x_4,\texttt{[M]},x_6\}$. Contiguously MLM is performed by minimizing the negative likelihood:
\begin{equation}
\resizebox{.89\linewidth}{!}{$
    \displaystyle
    -{\rm log}\ p_\theta(\bm{z}_{{\mathcal{M}}}|\bm{z}_{{\setminus\mathcal{M}}})=\!-\!\!\sum_{z\in \bm{z}_{\mathcal{M}}}\sum_{x\in z}{\rm log}\ p_\theta(x|\bm{z}_{{\setminus\mathcal{M}}}) .
$}
\end{equation}

\subsection{Explicitly N-gram Masked Language Modeling}
\label{sec3_3}
Different from contiguously MLM, we employ explicit $n$-gram identities as pre-training targets to reduce the prediction space for $n$-grams. To be specific, let $\bm{y}=\{y_1,...,y_{|\bm{b}|-1}\}, y\!\in\!\langle\mathcal{V}_F, \mathcal{V}_N\rangle$ to be the sequence of explicit $n$-gram identities,  $\bm{y}_{{\mathcal{M}}}$ to be the target $n$-gram identities, and $\bar{\bm{z}}_{{\setminus\mathcal{M}}}$ to be the sequence after explicitly masking $n$-grams. As shown in Figure~\ref{overview}(b), $\bm{y}_{\mathcal{M}}=\{y_2,y_4\},$ and $\bar{\bm{z}}_{{\setminus\mathcal{M}}}\!\!=\!\!\{x_1,\texttt{[M]},$ $x_4,\texttt{[M]},x_6\}$. For masked $n$-gram $\bm{x}_{[2:4)}$, the prediction space is significantly reduced from $\mathbb{R}^{|\mathcal{V}_F|^2}$ to $\mathbb{R}^{|\langle\mathcal{V}_F, \mathcal{V}_N\rangle|}$. Explicitly $n$-gram MLM is performed by minimizing the negative likelihood:
\begin{equation}
\setlength{\belowdisplayskip}{2pt}
\resizebox{.85\linewidth}{!}{$
    \displaystyle
    -{\rm log}\ p_\theta(\bm{y}_{{\mathcal{M}}}|\bar{\bm{z}}_{{\setminus\mathcal{M}}})=\!-\!\!\sum_{y\in \bm{y}_{\mathcal{M}}}{\rm log}\ p_\theta(y|\bar{\bm{z}}_{{\setminus\mathcal{M}}}) .
$}
\end{equation}

\begin{figure}[t]
\setlength{\belowcaptionskip}{-0.2cm}
\setlength{\abovecaptionskip}{4pt}
\begin{center} 
\includegraphics[width=0.87\linewidth]{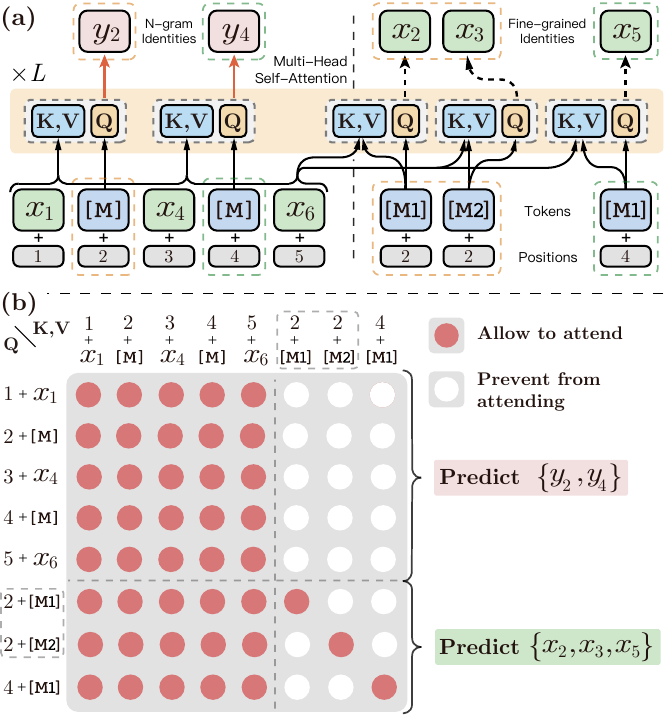}
\caption{(a) Detailed structure of Comprehensive N-gram MLM. (b) Self-attention mask $M$ without leaking length information of masked $n$-grams.}
\label{comprehensive_n-gram_mlm}
\end{center} 
\end{figure}

\subsection{Comprehensive N-gram Prediction}
\label{sec3_4}
We propose to simultaneously predict $n$-grams in fine-grained and coarse-grained manners corresponding to single mask symbol {\tt[M]}, which helps to extract comprehensive $n$-gram semantics, as shown in Figure~\ref{overview}(c). Comprehensive $n$-gram MLM is performed by minimizing the joint negative likelihood:
\begin{equation}
\setlength{\belowdisplayskip}{3pt}
\resizebox{1\linewidth}{!}{$
    \begin{split}
    &-{\rm log}\ p_\theta(\bm{y}_{{\mathcal{M}}},\bm{z}_{{\mathcal{M}}}|\bar{\bm{z}}_{{\setminus\mathcal{M}}})=\\
    &\!-\!\!\sum_{y\in \bm{y}_{\mathcal{M}}}{\rm log}\ p_\theta(y|\bar{\bm{z}}_{{\setminus\mathcal{M}}})
    -\!\sum_{z\in \bm{z}_{\mathcal{M}}}\sum_{x\in z}{\rm log}\ p_\theta(x|\bar{\bm{z}}_{{\setminus\mathcal{M}}}).
    \end{split}
$}
\end{equation}
where the predictions of explicit $n$-gram $\bm{y}_{\mathcal{M}}$ and fine-grained tokens $\bm{x}_{\mathcal{M}}$ are conditioned on the same context sequence $\bar{\bm{z}}_{{\setminus\mathcal{M}}}$.

In detail, to predict all tokens contained in a $n$-gram from single {\tt[M]} other than a consecutive sequence of {\tt[M]}, we adopt distinctive mask symbols {\tt[Mi]}$,\texttt{i}\!=\!1,...,n$ to aggregate contextualized representations for predicting the $\texttt{i}$-th token in $n$-gram. As shown in Figure~\ref{comprehensive_n-gram_mlm}(a), along with the same position as $y_2$, symbols {\tt[M1]} and {\tt[M2]} are used as queries ($Q$) to aggregate representations from $\bar{\bm{z}}_{{\setminus\mathcal{M}}}$ ($K$) for the predictions of $x_2$ and $x_3$, where $Q$ and $K$ donate the query and key in self-attention operation~\cite{transformer}. As shown in Figure~\ref{comprehensive_n-gram_mlm}(b),  the self-attention mask metric $M$ controls what context a token can attend to by modifying the attention weight $W_{A}\!=\!\texttt{softmax}(\frac{QK^T}{\sqrt{d_k}}+M)$, $M$ is assigned as:
\begin{equation}
\setlength{\abovedisplayskip}{7pt}
\setlength{\belowdisplayskip}{7pt}
\resizebox{.85\linewidth}{!}{$
    \displaystyle
M_{ij}=\begin{cases}
0,~~~~~~~ {\rm allow~to~attend}\\
-\infty,~~{\rm prevent~from~attending}
\end{cases}$}
\end{equation}

\begin{figure}[t]
\begin{center} 
\setlength{\belowcaptionskip}{-0.7cm}
\includegraphics[width=1\linewidth]{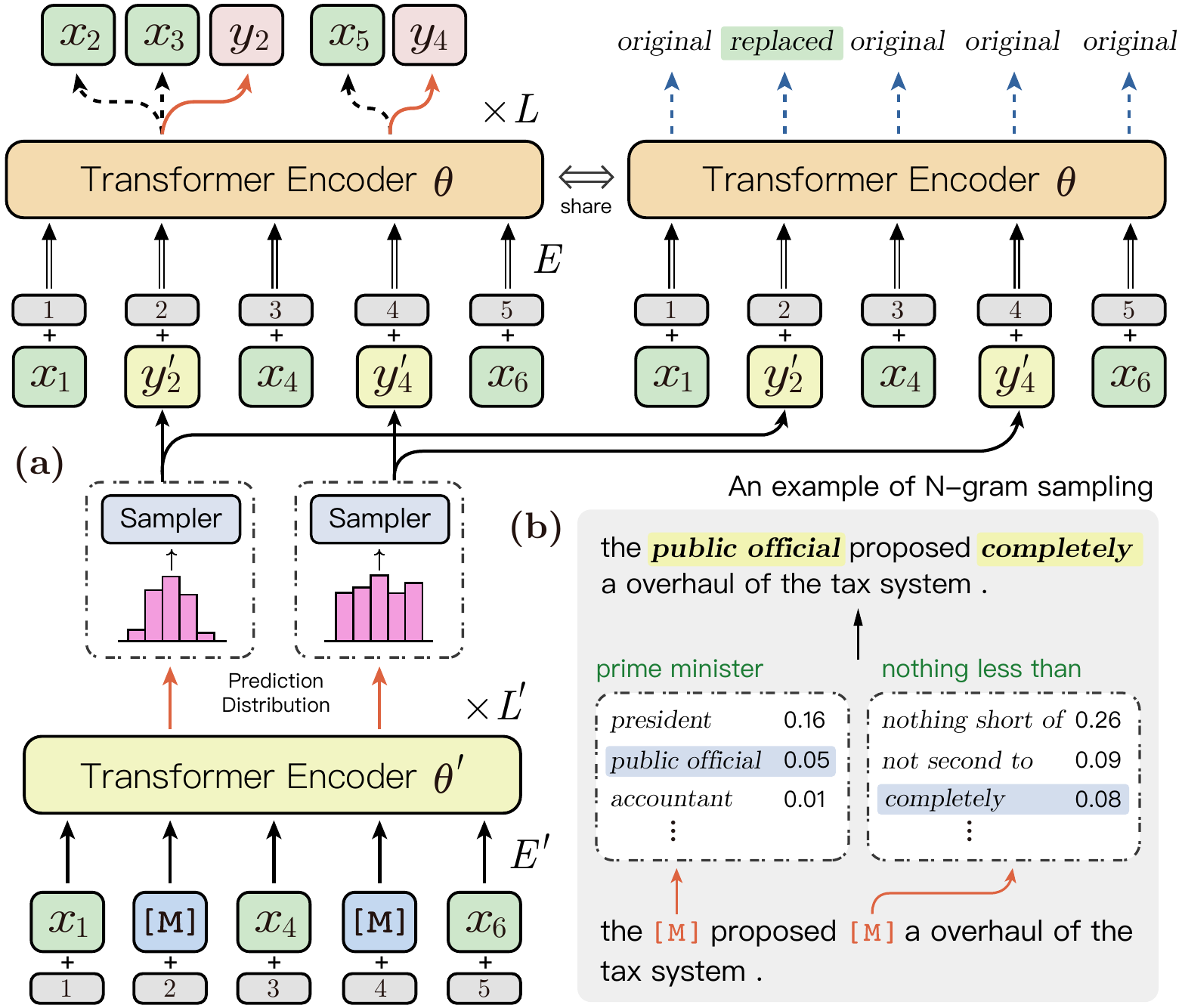}
\caption{(a) Detailed architecture of $n$-gram relation modeling, where $L^\prime$ donates the layers of the generator model. (b) An example of plausible $n$-gram sampling, where dotted boxes represent the sampling module, \textcolor{mygreen}{\textbf{texts}} in green are the original $n$-grams, and the \colorbox{myblue}{\textit{italic texts}} in blue donate the sampled $n$-grams.}
\label{n-gram_interaction}
\end{center} 
\end{figure}

We argue that the length information of $n$-grams is detrimental to the representations learning, because it will arbitrarily prune a number of semantically related $n$-grams with different lengths during predicting.  From this viewpoint, for the predictions of $n$-gram $\{x_2,x_3\}$, 1) we prevent context $\bar{\bm{z}}_{{\setminus\mathcal{M}}}$ from attending to $\{\!\texttt{[M1\!]},\!\texttt{[M2\!]}\!\}$ and 2) prevent $\{\!\texttt{[M1\!]},\!\texttt{[M2\!]}\!\}$ from attending to each other, so that the length information of $n$-grams will not be leaked in pre-training, as displayed in Figure~\ref{comprehensive_n-gram_mlm}(b).

\subsection{Enhanced N-gram Relation Modeling}
\label{sec3_5}
To explicitly learn the semantic relationships between $n$-grams, we jointly pre-train a small generator model $\theta^\prime$ with explicitly $n$-gram MLM objective to sample plausible $n$-gram identities. Then we employ the generated identities to preform masking and train the standard model $\theta$ to predict the original $n$-grams from fake ones in coarse-grained and fine-grained manners, as shown in Figure~\ref{n-gram_interaction}(a), which is efficient to model the pair relationships between similar $n$-grams. The generator model $\theta^\prime$ will not be used during fine-tuning, where the hidden size $H_{\theta^\prime}$ of $\theta^\prime$ has $H_{\theta^\prime}=H_{\theta}/3$ empirically.

As shown in Figure~\ref{n-gram_interaction}(b), $n$-grams of different length can be sampled to mask original $n$-grams according to the prediction distributions of $\theta^\prime$, which is more flexible and sufficient for constructing $n$-gram pairs than previous synonym masking methods~\cite{macbert} that require synonyms and original words to be of the same length. Note that our method needs a large embedding layer $E\in\mathbb{R}^{|\langle\mathcal{V}_F, \mathcal{V}_N\rangle|\times h}$ to obtain $n$-gram vectors in pre-training. To keep the number of parameters consistent with that of vanilla BERT, we remove the auxiliary embedding weights of $n$-grams during fine-tuning ($E\rightarrow E^\prime\in\mathbb{R}^{|\mathcal{V}_F|\times h}$). Specifically, let $\bm{y}^\prime_{{\mathcal{M}}}$ to be the generated $n$-gram identities, $\bar{\bm{z}}^\prime_{{\mathcal{M}}}$ to be the sequence masked by $\bm{y}^\prime_{{\mathcal{M}}}$, where $\bm{y}^\prime_{{\mathcal{M}}}=\{y^\prime_2, y^\prime_4\}$, and $\bar{\bm{z}}^\prime_{\setminus{\mathcal{M}}}=\!\{x_1,$ $y^\prime_2,x_4,y^\prime_4,x_6\}$ in Figure~\ref{n-gram_interaction}(a). The pre-training objective is to jointly minimize the negative likelihood of $\theta^\prime$ and $\theta$:
\begin{equation}
\resizebox{.89\linewidth}{!}{$
    \begin{split}
    -{\rm log}\ p_{\theta^\prime}(\bm{y}_{{\mathcal{M}}}|\bar{\bm{z}}_{{\setminus\mathcal{M}}})-{\rm log}\ p_{\theta}(\bm{y}_{{\mathcal{M}}},\bm{z}_{{\mathcal{M}}}|\bar{\bm{z}}^\prime_{{\setminus\mathcal{M}}}).
    \end{split}
$}
\end{equation}

Moreover, we incorporate the replaced token detection objective (RTD) to further distinguish fake $n$-grams from the mix-grained context $\bar{\bm{z}}^\prime_{{\setminus\mathcal{M}}}$ for interactions among explicit $n$-grams and fine-grained contextual tokens, as shown in the right part of Figure~\ref{n-gram_interaction}(a). Formally, we donate $\hat{\bm{z}}_{\setminus{\mathcal{M}}}$ to be the sequence after replacing masked $n$-grams with target $n$-gram identities $\bm{y}_{{\mathcal{M}}}$, the RTD objective is performed by minimizing the negative likelihood:
\begin{equation}
\setlength{\abovedisplayskip}{9pt}
\setlength{\belowdisplayskip}{4pt}
\resizebox{.89\linewidth}{!}{$
    \begin{split}
    &-{\rm log}\ p_{\theta}\big (\mathbbm{1}(\bar{\bm{z}}^\prime_{{\setminus\mathcal{M}}}=\hat{\bm{z}}_{\setminus{\mathcal{M}}})|\bar{\bm{z}}^\prime_{\setminus{\mathcal{M}}}\big )\\[-1.5mm]
    &=-\!\!\sum_{t=1}^{|\hat{\bm{z}}_{\setminus{\mathcal{M}}}|}{\rm log}\ p_{\theta}\big (\mathbbm{1}(\bar{\bm{z}}^\prime_{\setminus\mathcal{M},t}=\hat{\bm{z}}_{\setminus{\mathcal{M}},t})|\bar{\bm{z}}^\prime_{{\setminus\mathcal{M}}}, t\big ).
    \end{split}
$}
\end{equation}
As the example depicted in Figure~\ref{n-gram_interaction}(a), the target context sequence $\hat{\bm{z}}_{{\setminus\mathcal{M}}}=\{x_1,y_2,x_4,y_4,x_6\}$.
\subsection{N-gram Extraction}
\label{sec3_2}
\paragraph{N-gram Lexicon Extraction.} We employ T-test to extract semantically-complete $n$-grams statistically from unlabeled text corpora $\mathcal{X}$~\cite{ernie-gen}, as described in Algorithm~\ref{alg:n-gram}. 
\begin{algorithm}[h]
\caption{N-gram Extraction with T-test}
\label{alg:n-gram}
\vspace{1pt}
{\fontsize{10pt}{0}\selectfont\textbf{Input:} Large-scale text corpora $\mathcal{X}$ for pre-training}\\
{\fontsize{10pt}{0}\selectfont\textbf{Output:} Semantic $n$-gram lexicon $\mathcal{V}_N$}\\
\hfill\AlgCommentInLine{{\fontsize{9pt}{0}\selectfont {\rm {\fontsize{10pt}{0}\selectfont given initial hypothesis $H_0$: a randomly constructed $n$-gram $\bm{w}\!\!=\!\!\{x_1,...,x_n\}$ with probability $p^\prime(\bm{w})\!=\!\prod_{i=1}^n p(x_{i})$ cannot be a statistically semantic $n$-gram}}}}\\
\For{{\fontsize{10pt}{0}\selectfont l {\rm in} range{\rm (2, $n$)}}}{{
\fontsize{10pt}{0}\selectfont $\mathcal{V}_{N_l}\leftarrow\langle\rangle$\hfill\AlgCommentInLine{{\fontsize{9pt}{0}\selectfont {\rm initialize the lexicon for $l$-grams}}}}\\
\For{{\fontsize{10pt}{0}\selectfont l{\rm -gram} $\bm{w}$ {\rm in}  $\mathcal{X}$}} {{\fontsize{10pt}{0}\selectfont
    $s\leftarrow{\frac{(p(\bm{w})-p^\prime(\bm{w}))}{\sqrt{\sigma^2/N_l}}}$: $t$-statistic score\vspace{-3pt}\hfill\AlgCommentInLine{{\fontsize{9pt}{0}\selectfont {\rm where statistical probability $p(\bm{w})=\frac{\texttt{Count}(\bm{w})}{N_l}$, deviation $\sigma^2\!=p(\bm{w})(1-p(\bm{w}))$, $N_l$ donates the count of $l$-grams in $\mathcal{X}$}}}\\$\mathcal{V}_{N_l}.$\textit{append}$(\{\bm{w}, s\})$}}
    {\fontsize{10pt}{0}\selectfont $\mathcal{V}_{N_l}\leftarrow$\;\textit{topk}$(\mathcal{V}_{N_l}, k_l)$\hfill\AlgCommentInLine{{\fontsize{9pt}{0}\selectfont {\rm $k_l$ is the number of $l$-gram}}}}
    }
    {\fontsize{10pt}{0}\selectfont $\mathcal{V}_N\leftarrow\langle\mathcal{V}_{N_2},...,\mathcal{V}_{N_n}\rangle$}\hfill\AlgCommentInLine{{\fontsize{9pt}{0}\selectfont {\rm merge all lexicons}}}\\
\Return{{\fontsize{10pt}{0}\selectfont $\mathcal{V}_N$}}
\end{algorithm}
We first calculate the $t$-statistic scores of all $n$-grams appearing in $\mathcal{X}$ since the higher the $t$-statistic score, the more likely it is a semantically-complete $n$-gram. Then, we select the $l$-grams with the top $k_l$ $t$-statistic scores to construct the final $n$-gram lexicon $\mathcal{V}_{N}$\vspace{-5pt}. 

\paragraph{N-gram Boundary Extraction.} To incorporate $n$-gram information into MLM objective, $n$-gram boundaries are referred to mask whole $n$-grams for pre-training. Given an input sequence $\bm{x}=\{x_1,...,x_{|\bm{x}|}\}$, we employ maximum matching algorithm to traverse valid $n$-gram paths $\mathcal{B}=\{\bm{b}_1,...,\bm{b}_{|\mathcal{B}|}\}$ according to $\mathcal{V}_{N}$, then select the shortest paths as the final $n$-gram boundaries $\bm{b}$, where $|\bm{b}|\leq|\bm{b}_i|, \forall i=1,...,|\mathcal{B}|$.

\begin{table*}[t]
\setlength{\belowcaptionskip}{-0.3cm}
\centering
\renewcommand\tabcolsep{4pt}
\resizebox{0.95\textwidth}{!}{
\begin{tabular}{l|c|cccccccc|c}
\toprule
\multirow{2}{*}{\textbf{Models}} & \multirow{2}{*}{\textbf{\#Param}} & \textbf{MNLI} & \textbf{QNLI} & \textbf{QQP} & \textbf{SST-2} & \textbf{CoLA} & \textbf{MRPC} & \textbf{RTE} & \textbf{STS-B} & \textbf{GLUE} \\ 
&&Acc&Acc&Acc&Acc&MCC&Acc&Acc&PCC&\textbf{Avg} \\
\midrule
\multicolumn{11}{l}{\textit{Results of single models pre-trained on \textbf{base-scale} text corpora (16GB)}} \\
\midrule
BERT~\cite{bert}&110M&84.5&91.7&91.3&93.2&58.9&87.3&68.6&89.5&83.1 \\
TUPE~\cite{tupe}&110M&86.2&92.1&91.3&\textbf{93.3}&63.6&89.9&73.6&89.2&85.0 \\
F-TFM$_{{\rm ELECTRA}}$~\cite{f-tfm}&110M&86.4&92.1&91.7&93.1&64.3&89.2&75.4&\textbf{90.8}&85.4\vspace{-6pt} \\
\multicolumn{11}{l}{\dashrule\vspace{-3pt}} \\
ERNIE-Gram&110M&\textbf{87.1}&\textbf{92.8}&\textbf{91.8}&93.2&\textbf{68.5}&\textbf{90.3}&\textbf{79.4}&90.4&\textbf{86.7}\\
\midrule
\multicolumn{11}{l}{\textit{Results of single models pre-trained on \textbf{large-scale} text corpora (160GB or more)}} \\
\midrule
XLNet~\cite{xlnet}&110M&86.8&91.7&91.4&94.7&60.2&88.2&74.0&89.5&84.5\\
RoBERTa~\cite{roberta}&135M&87.6&92.8&91.9&94.8&63.6&90.2&78.7&91.2&86.4\\
ELECTRA~\cite{electra}&110M&88.8&93.2&91.5&95.2&67.7&89.5&82.7&91.2&87.5\\
\textsc{UniLM}v2~\cite{unilm}&110M&88.5&\textbf{93.5}&91.7&95.1&65.2&\textbf{91.8}&81.3&91.0&87.3\\
MPNet~\cite{mpnet}&110M&88.5&93.3&91.9&95.4&65.0&91.5&\textbf{85.2}&90.9&87.7\vspace{-6pt}\\
\multicolumn{11}{l}{\dashrule\vspace{-3pt}} \\
ERNIE-Gram&110M&\textbf{89.1}&93.2&\textbf{92.2}&\textbf{95.6}&\textbf{68.6}&90.7&83.8&\textbf{91.3}&\textbf{88.1}\\

\bottomrule
\end{tabular}
}
\caption{Results on the development set of the GLUE benchmark for base-size pre-trained models. Models using 16GB corpora are all pre-trained with a batch size of 256 sequences for 1M steps. STS-B and CoLA are reported by Pearson correlation coefficient (PCC) and Matthews correlation coefficient (MCC), other tasks are reported by accuracy (Acc). Note that results of ERNIE-Gram are the median of over ten runs with different random seeds.}
\label{table:glue}
\end{table*}

\section{Experiments}
\label{sec:ex}
In this section, we first present the pre-training configuration of ERNIE-Gram on Chinese and English text corpora. 
Then we compare ERNIE-Gram with previous works on various downstream tasks. We also conduct several ablation experiments to access the major components of ERNIE-Gram.
\subsection{Pre-training Text Corpora}
\paragraph{English Pre-training Data.} We use two common text corpora for English pre-training:
\begin{itemize}[leftmargin=*]
\vspace{-7pt}
\item \textbf{Base-scale corpora:} 16GB uncompressed text from \textsc{Wikipedia} and \textsc{BooksCorpus}~\cite{bookcorpus}, which is the original data for BERT.
\vspace{-20pt}
\item \textbf{Large-scale corpora:} 160GB uncompressed text from \textsc{Wikipedia}, \textsc{BooksCorpus}, \textsc{OpenWebText}\footnote{\url{http://web.archive.org/save/http://Skylion007.github.io/OpenWebTextCorpus}}, \textsc{CC-News}~\cite{roberta} and \textsc{Stories}~\cite{stories}, which is the original data used in RoBERTa.
\vspace{-7pt}
\end{itemize}
\paragraph{Chinese Pre-training Data.} We adopt the same Chinese text corpora used in ERNIE2.0~\cite{ernie2} to pre-train ERNIE-Gram.
\subsection{Pre-training Setup}
Before pre-training, we first extract $200$K bi-grams and $100$K tri-grams with Algorithm~\ref{alg:n-gram} to construct the semantic $n$-gram lexicon $\mathcal{V}_N$ for English and Chinese corpora. and we adopt the sub-word dictionary ($30$K BPE codes) used in BERT  and the character dictionary used in ERNIE2.0 as our fine-grained vocabulary $\mathcal{V}_F$ in English and Chinese. 

Following the previous practice, we pre-train ERNIE-Gram in base size ($L=12, H=768,$ $A=12$, Total Parameters=$110$M)\footnote{We donate the number of layers as $L$, the hidden size as $H$ and the number of self-attention heads as $A$.}, and set the length of the sequence in each batch up to $512$ tokens. We add the relative position bias~\cite{t5} to attention weights and use Adam~\cite{adam} for optimizing. For pre-training on base-scale English corpora, the batch size is set to $256$ sequences, the peak learning rate is $1e$-$4$ for $1$M training steps, which are the same settings as BERT$_{{\rm BASE}}$. As for large-scale English corpora, the batch size is $5112$ sequences, the peak learning rate is $4e$-$4$ for $500$K training steps. For pre-training on Chinese corpora, the batch size is $256$ sequences, the peak learning rate is $1e$-$4$ for 3M training steps. All the pre-training hyper-parameters are supplemented in the Appendix~\ref{appendix_a}. 

In fine-tuning, we remove the auxiliary embedding weights of explicit $n$-grams identities for fair comparison with previous pre-trained models.

\subsection{Results on GLUE Benchmark}
The General Language Understanding Evaluation (GLUE;~\citealp{glue}) is a multi-task benchmark consisting of various NLU tasks, which contains 1) pairwise classification tasks like language inference (MNLI;~\citealp{mnli}, RTE;~\citealp{rte}), question answering (QNLI;~\citealp{squad1}) and paraphrase detection (QQP, MRPC;~\citealp{mrpc2005}), 2) single-sentence classification tasks like linguistic acceptability (CoLA;~\citealp{cola2018}), sentiment
analysis (SST-2;~\citealp{sst2013}) and 3) text similarity task (STS-B;~\citealp{sts-b2017}). 

The fine-tuning results on GLUE of ERNIE-Gram and various strong baselines are presented in Table~\ref{table:glue}. For fair comparison, the listed models are all in base size and fine-tuned without any data augmentation. Pre-trained with base-scale text corpora, ERNIE-Gram outperforms recent models such as TUPE and F-TFM by $1.7$ and $1.3$ points on average. As for large-scale text corpora, ERNIE-Gram achieves average score increase of $1.7$ and $0.6$ over RoBERTa and ELECTRA, demonstrating the effectiveness of ERNIE-Gram.

\begin{table}[t]
\centering
\setlength{\belowcaptionskip}{-0.2cm}
\setlength{\belowcaptionskip}{-0.2cm}
\resizebox{0.49\textwidth}{!}{
\begin{tabular}{l|cccc}
\toprule
\multirow{2}{*}{\textbf{Models}} & \multicolumn{2}{c}{\textbf{SQuAD1.1}} & \multicolumn{2}{c}{\textbf{SQuAD2.0}} \\ 
&EM&F1&EM&F1\\
\midrule
\multicolumn{5}{l}{\textit{Models pre-trained on \textbf{base-scale} text corpora (16GB)}} \\
\midrule
BERT~\cite{bert}& 80.8 & 88.5 & 73.7 & 76.3 \\
RoBERTa~\cite{roberta}& - & 90.6 & - & 79.7 \\
XLNet~\cite{xlnet}& - & - & 78.2 & 81.0 \\
MPNet~\cite{mpnet}& 85.0 & 91.4 & 80.5 & 83.3 \\
\textsc{UniLM}v2~\cite{unilm}& 85.6 & 92.0 & 80.9 & 83.6\vspace{-8pt} \\
\multicolumn{5}{l}{\dashrule\vspace{-3pt}} \\
ERNIE-Gram& \textbf{86.2} & \textbf{92.3} & \textbf{82.1} & \textbf{84.8}\\
\midrule
\multicolumn{5}{l}{\textit{Models pre-trained on \textbf{large-scale} text corpora (160GB)}} \\
\midrule
RoBERTa~\cite{roberta}& 84.6 & 91.5 & 80.5 & 83.7 \\
XLNet~\cite{xlnet}& - & - & 80.2 & - \\
ELECTRA~\cite{electra}& 86.8 & - & 80.5 & -\\
MPNet~\cite{mpnet}& 86.8 & 92.5 & 82.8 & 85.6 \\
\textsc{UniLM}v2~\cite{unilm}& 87.1 & 93.1 & 83.3 & 86.1\vspace{-8pt} \\
\multicolumn{5}{l}{\dashrule\vspace{-3pt}} \\
ERNIE-Gram& \textbf{87.2} & \textbf{93.2} & \textbf{84.1} & \textbf{87.1}\\
\bottomrule
\end{tabular}
}
\caption{Performance comparison between base-size pre-trained models on the SQuAD development sets. Exact-Match (EM) and F1 score are adopted for evaluations. Results of ERNIE-Gram are the median of over ten runs with different random seeds.}
\label{table:squad}
\end{table}

\subsection{Results on Question Answering (SQuAD)}
The Stanford Question Answering (SQuAD) tasks are designed to extract the answer span within the given passage conditioned on the question. We conduct experiments on SQuAD1.1 ~\cite{squad1} and SQuAD2.0 ~\cite{squad2} by adding a classification layer on the sequence outputs of ERNIE-Gram and predicting whether each token is the start or end position of the answer span. 
Table~\ref{table:squad} presents the results on SQuAD for base-size pre-trained models, ERNIE-Gram achieves better performance than current strong baselines on both base-scale and large-scale pre-training text corpora\vspace{-5pt}.

\begin{table}[t]
\setlength{\belowcaptionskip}{-0.3cm}
\centering
\resizebox{0.48\textwidth}{!}{
\begin{tabular}{l|ccccc}
\toprule
\multirow{2}{*}{\textbf{Models}} &\multicolumn{3}{c}{\textbf{RACE}} & \textbf{IMDb} & \textbf{AG} \\ 
&Total&High&Middle&Err.&Err.\\
\midrule
\multicolumn{6}{l}{\textit{Pre-trained on \textbf{base-scale} text corpora (16GB)}} \\
\midrule
BERT$^{\;a}$&65.0&62.3&71.7& 5.4 & 5.9   \\
XLNet$^{\;b}$& 66.8&-&-&4.9 & -  \\
MPNet$^{\;c}$&70.4&67.7&76.8& 4.8 & -  \\
F-TFM$^{\ d}_{{\rm ELECTRA}}$ &-&-&-& 5.2 & 5.4\vspace{-6pt} \\
\multicolumn{6}{l}{\dashrule\vspace{-3pt}} \\
ERNIE-Gram& \textbf{72.7}&\textbf{68.1}&75.1&\textbf{4.6} & \textbf{5.0} \\
\midrule
\multicolumn{6}{l}{\textit{Pre-trained on \textbf{large-scale} text corpora (160GB)}} \\
\midrule
MPNet$^{\;c}$&72.0&70.3&76.3& 4.4 & - \vspace{-8pt} \\
\multicolumn{6}{l}{\dashrule\vspace{-3pt}} \\
ERNIE-Gram& \textbf{77.7}&\textbf{75.6}&\textbf{78.8}&\textbf{3.9} & \textbf{4.9} \\
\bottomrule
\end{tabular}
}
\caption{Comparison on the test sets of RACE, IMDb and AG. The listed models are all in base-size. In the results of RACE, ``High" and ``Middle" represent the training and evaluation sets for high schools and middle schools respectively, ``Total" is the full training and evaluation set. $^a$~\cite{bert}; $^b$~\cite{xlnet}; $^c$~\cite{mpnet}; $^d$~\cite{f-tfm}.}
\label{table:imdb}
\end{table}

\begin{table*}
\setlength{\belowcaptionskip}{-0.3cm}
\centering
\resizebox{0.99\textwidth}{!}{
\renewcommand\tabcolsep{2.8pt}
\begin{tabular}{l|cccccccccc}
\toprule
\multirow{3}{*}{\textbf{Models}} & \multicolumn{2}{c}{\textbf{XNLI}} & \multicolumn{2}{c}{\textbf{LCQMC}} & \multicolumn{2}{c}{\textbf{DRCD}} &\textbf{CMRC2018}&\textbf{DuReader}& \multicolumn{2}{c}{\textbf{M-NER}} \\ 
& \multicolumn{2}{c}{Acc} & \multicolumn{2}{c}{Acc} & \multicolumn{2}{c}{EM / F1} & EM / F1& EM / F1 & \multicolumn{2}{c}{F1}\\ 
&Dev&Test&Dev&Test&Dev&Test&Dev&Dev&Dev&Test\\
\midrule 
RoBERTa-wwn-ext$_{{\rm LARGE}}^\ast$& 82.1 &81.2 &90.4&87.0&89.6 / 94.8&89.6 / 94.5&68.5 / 88.4&- / -&-&-\\
NEZHA$_{{\rm LARGE}}$~\cite{nezha}& 82.2 &81.2 &90.9&87.9&- / -&- / -&- / -&- / -&-&-\\
MacBERT$_{{\rm LARGE}}$~\cite{macbert}& 82.4 &81.3 &90.6&87.6&91.2 / 95.6&91.7 / 95.6&70.7 / 88.9&- / -&-&-\\
\midrule 
BERT-wwn-ext$^\ast_{{\rm BASE}}$& 79.4 &78.7 &89.6&87.1&85.0 / 91.2&83.6 / 90.4&67.1 / 85.7&- / -&-&-\\
RoBERTa-wwn-ext$^\ast_{{\rm BASE}}$& 80.0 &78.8 &89.0&86.4&85.6 / 92.0&67.4 / 87.2&67.4 / 87.2&- / -&-&-\\
\textsc{Zen}$_{{\rm BASE}}$~\cite{zen}& 80.5 &79.2 &90.2&88.0&- / -&- / -&- / -&- / -&-&-\\
NEZHA$_{{\rm BASE}}$~\cite{nezha}& 81.4 &79.3 &90.0&87.4&- / -&- / -&- / -&- / -&-&-\\
MacBERT$_{{\rm BASE}}$~\cite{macbert}& 79.0 &78.2 &89.4&87.0&88.3 / 93.5&87.9 / 93.2&69.5 / 87.7&- / -&-&-\\
ERNIE1.0$_{{\rm BASE}}$~\cite{ernie1}& 79.9 &78.4 &89.7&87.4&84.6 / 90.9&84.0 / 90.5&65.1 / 85.1&57.9 / 72.1&95.0&93.8\\
ERNIE2.0$_{{\rm BASE}}$~\cite{ernie2}& 81.2 &79.7 &\textbf{90.9}&87.9&88.5 / 93.8&88.0 / 93.4&69.1 / 88.6&61.3 / 74.9&95.2&93.8\vspace{-6pt} \\
\multicolumn{11}{l}{\dashrule\vspace{-3pt}}\\
ERNIE-Gram$_{{\rm BASE}}$& \textbf{81.8} &\textbf{81.5} &90.6&\textbf{88.5}&\textbf{90.2} / \textbf{95.0}&\textbf{89.9} / \textbf{94.6}&\textbf{74.3} / \textbf{90.5}&\textbf{64.2} / \textbf{76.8}&\textbf{96.5}&\textbf{95.3}\\
\bottomrule
\end{tabular}
}
\caption{Results on six Chinese NLU tasks for base-size pre-trained models. Results of models with asterisks ``$^\ast$" are from ~\citealp{bert-wwm}. M-NER is in short for MSRA-NER dataset. ``${{\rm BASE}}$" and ``${{\rm LARGE}}$" donate different sizes of pre-training models. Large size models have $L=24, H=1024, A=16$ and total Parameters=$340$M.}
\label{table:chinese1}
\end{table*}

\subsection{Results on RACE and Text Classification Tasks}
The ReAding Comprehension from Examinations (RACE;~\citealp{race}) dataset collects $88$K long passages from English exams at middle and high schools, the task is to select the correct choice from four given options according to the questions and passages. We also evaluate ERNIE-Gram on two large scaled text classification tasks that involve long text and reasoning, including sentiment analysis datasets IMDb ~\cite{imdb} and topic classification dataset AG's News ~\cite{ag}. The results are reported in Table~\ref{table:imdb}. It can be seen that ERNIE-Gram consistently outperforms previous models, showing the advantage of ERNIE-Gram on tasks involving long text and reasoning.

\subsection{Results on Chinese NLU Tasks}
We execute extensive experiments on six Chinese language understanding tasks, including natural language inference (XNLI;~\citealp{xnli}), machine reading comprehension (CMRC2018;~\citealp{DBLP:journals/corr/abs-1810-07366}, DRCD;~\citealp{drcd} and DuR-eader;~\citealp{dureader}), named entity recognition (MSRA-NER;~\citealp{ner}) and semantic similarity (LCQMC;~\citealp{lcqmc}).

Results on six Chinese tasks are presented in Table~\ref{table:chinese1}. It is observed that ERNIE-Gram significantly outperforms previous models across tasks by a large margin and achieves new state-of-the-art results on these Chinese NLU tasks in base-size model group. Besides, ERNIE-Gram$_{{\rm BASE}}$ are also better than various large-size models on XNLI, LCQMC and CMRC2018 datasets.
\subsection{Ablation Studies}
\begin{figure}
\begin{center} 
\setlength{\belowcaptionskip}{-0.4cm}
\includegraphics[width=1\linewidth]{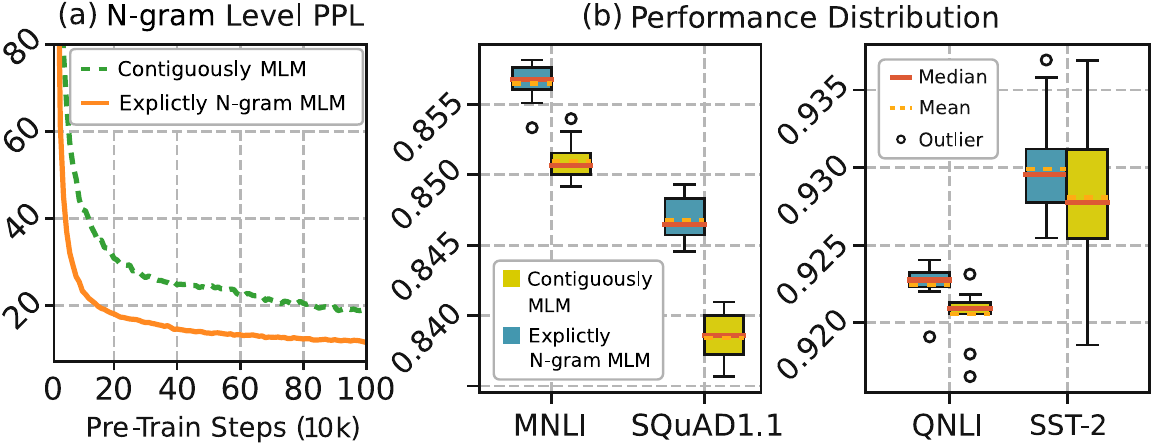}
\caption{(a) N-gram level perplexity which is calculated by $ (\prod_{i=1}^{k}\texttt{PPL}({\bm w}_i))^{\frac{1}{k}}$ for contiguously MLM, where ${\bm w}_i$ is the $i$-th masked $n$-gram. (b) Performance distribution box plot on MNLI, QNLI, SST-2 and SQuAD1.1.}
\label{EMLM}
\end{center}
\end{figure}

We further conduct ablation experiments to analyze the major components of ERNIE-Gram.\vspace{-0.1cm}
\paragraph{Effect of Explicitly N-gram MLM.}We compare two models pre-trained with contiguously MLM and explicitly $n$-gram MLM objectives in the same settings (the size of $n$-gram lexicon is $300$K). The evaluation results for pre-training and fine-tuning are shown in Figure~\ref{EMLM}. Compared with contiguously MLM, explicitly $n$-gram MLM objective facilitates the learning of $n$-gram semantic information with lower $n$-gram level perplexity in pre-training and better performance on downstream tasks. This verifies the effectiveness of explicitly $n$-gram MLM objective for injecting $n$-gram semantic information into pre-training.\vspace{-0.1cm}

\begin{table*}[t]
\small
\centering
\renewcommand\tabcolsep{5pt}
\begin{minipage}[c]{0.635\linewidth}

\begin{tabular}{cl|ccccccc}
\toprule
\multirow{2}{*}{$\texttt{\bm{\#}}$}&\multirow{2}{*}{\textbf{Models}}  & \multicolumn{2}{c}{\textbf{MNLI}}  & \textbf{SST-2} & \multicolumn{2}{c}{\textbf{SQuAD1.1}} & \multicolumn{2}{c}{\textbf{SQuAD2.0}}  \\ 
&&m&mm&Acc&EM&F1&EM&F1 \\
\midrule
&XLNet$^{\;a}$&85.6&85.1&93.4&-&-&78.2&81.0 \\
&RoBERTa$^{\;b}$&84.7&-&92.7&-&90.6&-&79.7 \\
&MPNet$^{\;c}$&85.6&-&\textbf{93.6}&84.0&90.3&79.5&82.2\\
&\textsc{UniLM}v2$^{\;d}$&85.6&85.5&93.0&85.0&91.5&78.9&81.8\\
\midrule
$\texttt{\#1}$&ERNIE-Gram&\textbf{86.5}&\textbf{86.4}&93.2&\textbf{85.2}&\textbf{91.7}&\textbf{80.8}&\textbf{84.0}\\
$\texttt{\#2}$&$-\ ${\fontsize{10pt}{0}\selectfont {\tt CNP}}&86.2&86.2&92.7&85.0&91.5&80.4&83.4\\
$\texttt{\#3}$&$-\ ${\fontsize{10pt}{0}\selectfont {\tt ENRM}}&85.7&85.8&93.5&84.7&91.3&79.7&82.7\\
$\texttt{\#4}$&$-\ ${\fontsize{10pt}{0}\selectfont {\tt CNP}}$\ -\ ${\fontsize{10pt}{0}\selectfont {\tt ENRM}}&85.6&85.7&92.9&84.5&91.2&79.5&82.4\\
\bottomrule
\end{tabular}
\setlength{\belowcaptionskip}{-0.3cm}
\caption{Comparisons between comprehensive $n$-gram prediction ({\fontsize{10pt}{0}\selectfont {\tt CNP}}) and enhanced $n$-gram relation modeling ({\fontsize{10pt}{0}\selectfont {\tt ENRM}}) methods. All the listed models are pre-trained following the same settings of BERT$_{{\rm BASE}}$~\cite{bert} and without relative position bias. Results of ERNIE-Gram variants are the median of over ten runs with different random seeds. Results in the upper block are from $^a$~\cite{xlnet}, $^b$~\cite{roberta}, $^c$~\cite{mpnet} and $^d$~\cite{unilm}.}
\label{table:ablation}
\end{minipage}
\hfill
\begin{minipage}[c]{.35\linewidth}
\begin{center} 
\includegraphics[width=1\linewidth]{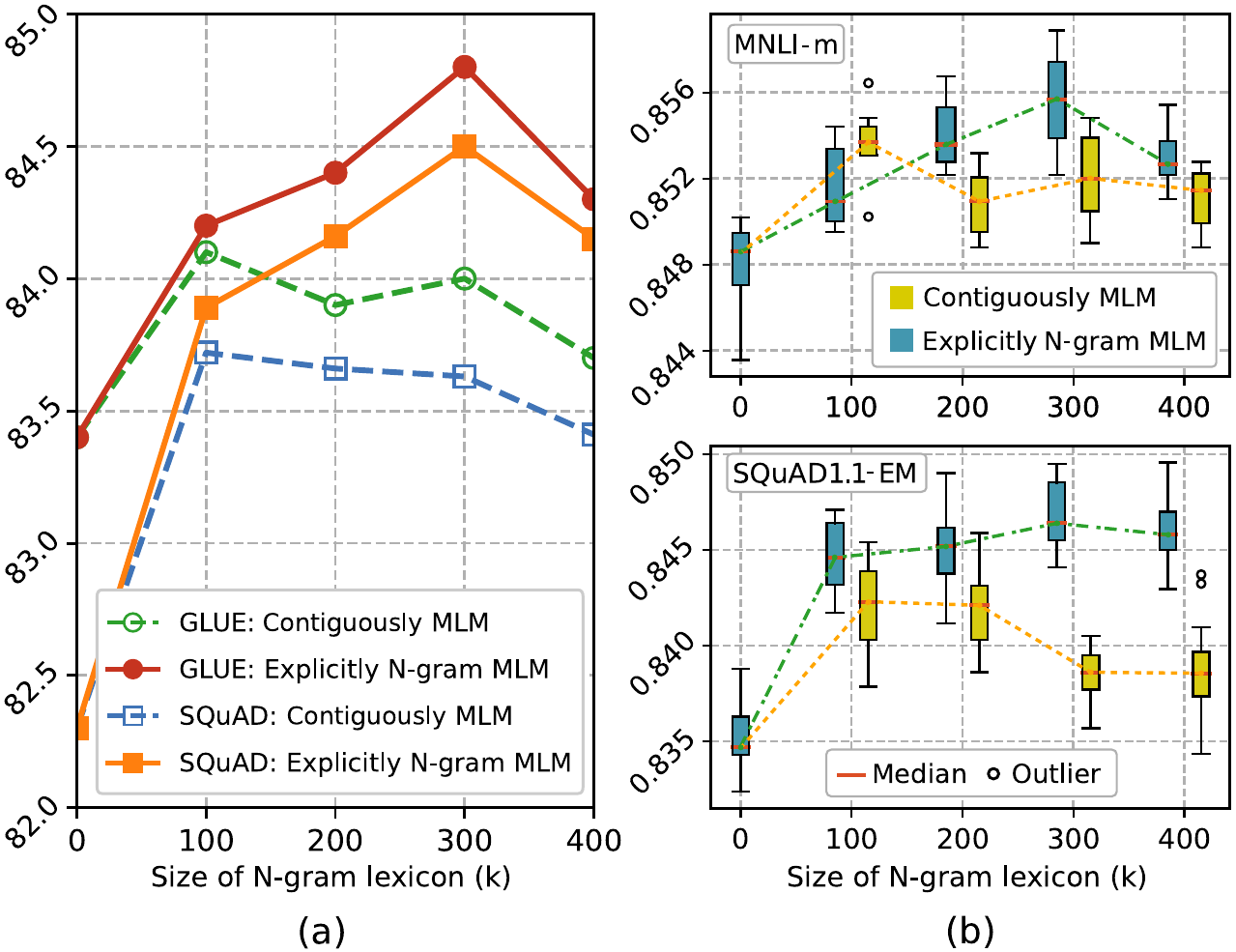}
\captionsetup{type=figure}
\setlength{\abovecaptionskip}{-0.11cm}
\setlength{\belowcaptionskip}{-0.3cm}
\caption{Quantitative study on the size of extracted $n$-gram lexicon. (a) Comparisons on GLUE and SQuAD. Note that SQuAD is presented by the average scores of SQuAD1.1 and SQuAD2.0. (b) Performance distribution box plot on MNLI and SQuAD1.1 datasets.}
\label{size_fig}
\end{center} 
\end{minipage}
\end{table*}

\paragraph{Size of N-gram Lexicon.} To study the impact of $n$-gram lexicon size on model performance, we extract $n$-gram lexicons with size from $100$K to $400$K for pre-training, as shown in Figure \ref{size_fig}. As the lexicon size enlarges, performance of contiguously MLM becomes worse, presumably because more $n$-grams are matched and connected as longer consecutive spans for prediction, which is more difficult for representation learning. Explicitly $n$-gram MLM with lexicon size being $300$K achieves the best results, while the performance significantly declines when the size of lexicon increasing to $400$K because more low-frequent $n$-grams are learning unnecessarily. See Appendix~\ref{appedix_c} for detailed results of different lexicon choices on GLUE and SQuAD.\vspace{-0.2cm}

\paragraph{Effect of Comprehensive N-gram Prediction and Enhanced N-gram Relation Modeling.} As shown in Table~\ref{table:ablation}, we compare several ERNIE-Gram variants with previous strong baselines under the BERT$_{{\rm BASE}}$ setting. After removing comprehensive $n$-gram prediction ($\texttt{\#2}$), ERNIE-Gram degenerates to a variant with explicitly $n$-gram MLM and $n$-gram relation modeling and its performance drops slightly by $0.3$-$0.6$. When removing enhanced $n$-gram relation modeling ($\texttt{\#3}$), ERNIE-Gram degenerates to a variant with comprehensive $n$-gram MLM and the performance drops by $0.4$-$1.3$. If removing both comprehensive $n$-gram prediction and relation modeling ($\texttt{\#4}$), ERNIE-Gram degenerates to a variant with explicitly $n$-gram MLM and the performance drops by $0.7$-$1.6$. These results demonstrate the advantage of comprehensive $n$-gram prediction and $n$-gram relation modeling methods for efficiently $n$-gram semantic injecting into pre-training. The detailed results of ablation study are supplemented in Appendix~\ref{appedix_c}.

\begin{figure}[t]
\begin{center} 
\setlength{\abovecaptionskip}{-0.0cm}
\setlength{\belowcaptionskip}{-0.6cm}
\includegraphics[width=1\linewidth]{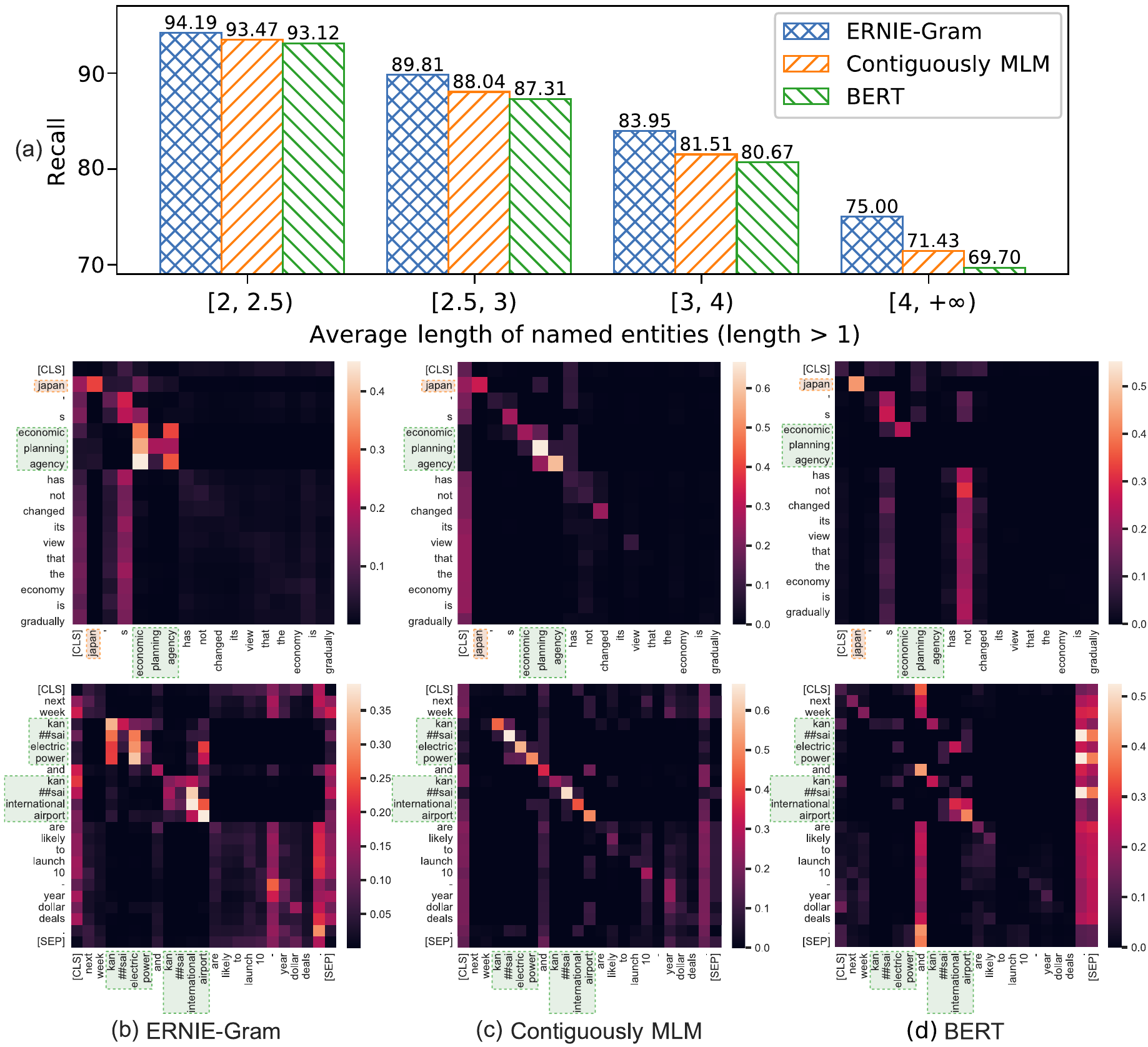}
\caption{(a) Recall rate of whole named entities on different evaluation subsets, which have incremental average length of named entities. (b-d) Mean attention scores of 12 attention heads in the last self-attention layer. Texts in \colorbox{mygreen1}{green} and \colorbox{myorange1}{orange} boxes are named entities standing for organizations and locations.}
\label{case}
\end{center}
\end{figure}

\subsection{Case Studies\vspace{-2pt}}
To further understand the effectiveness of our approach for learning $n$-grams information, we fine-tune ERNIE-Gram, contiguously MLM and lower-cased BERT on CoNLL-2003 named entity recognition task~\cite{conll03} for comparison.  
We divide the evaluation set into five subsets based on the average length of the named entities in each sentence. As shown in Figure~\ref{case}(a), it is more difficult to recognize whole named entities as the length of them increases, while the performance of ERNIE-Gram declines slower than contiguously MLM and BERT, which implies that ERNIE-Gram models tighter intra-dependencies of $n$-grams.

As shown in Figure~\ref{case}(b-d), we visualize the attention patterns in the last self-attention layer of fine-tuned models. For contiguously MLM, there are clear diagonal lines in named entities that tokens prefer to attend to themself in named entities. While for ERNIE-Gram, there are bright blocks over named entities that tokens attend to most of tokens in the same entity adequately to construct tight representation, verifying the effectiveness of ERNIE-Gram for $n$-gram semantic modeling.

\section{Conclusion}
In this paper, we present ERNIE-Gram, an explicitly $n$-gram masking and predicting method to eliminate the limitations of previous contiguously masking strategies and incorporate coarse-grained linguistic information into pre-training sufficiently. ERNIE-Gram conducts comprehensive $n$-gram prediction and relation modeling to further enhance the learning of semantic $n$-grams for pre-training. Experimental results on various NLU tasks demonstrate that ERNIE-Gram outperforms XLNet and RoBERTa by a large margin, and achieves state-of-the-art results on various benchmarks. Future work includes constructing more comprehensive $n$-gram lexicon ($n\!>\!3$) and pre-training ERNIE-Gram with large-size model for more downstream tasks.

\section*{Acknowledgments}
We would like to thank Zhen Li for his constructive suggestions, and hope everything goes well with his work. We are also indebted to the NAACL-HLT reviewers for their detailed and insightful comments on our work.\vspace{-5pt}

\bibliography{ernie-gram}
\bibliographystyle{acl_natbib}

\clearpage
\appendix
\appendix
\section{Hyperparameters for Pre-Training}
\label{appendix_a}
As shown in Table~\ref{table:pt-hparam}, we list the detailed hyperparameters used for pre-training ERNIE-Gram on base and large scaled English text corpora and Chinese text corpora. We follow the same hyperparameters of BERT$_{{\rm BASE}}$~\cite{bert} to pre-train ERNIE-Gram on the base-scale English text corpora (16GB). We pre-train ERNIE-Gram on the large-scale text corpora (160GB) with the settings in RoBERTa~\cite{roberta} except the batch size being 5112 sequences. \vspace{-2pt}

\begin{table}[H]
\centering
\setlength{\belowcaptionskip}{-0.5cm}
\small
\resizebox{0.48\textwidth}{!}{
\renewcommand\tabcolsep{2.1pt}
\begin{tabular}{lccc}
\toprule
\textbf{Hyperparameters} & \textbf{Base-scale} & \textbf{Large-scale} & \textbf{Chinese} \\ \midrule
Layers & \multicolumn{3}{c}{12}\\
Hidden size & \multicolumn{3}{c}{768} \\
Attention heads &\multicolumn{3}{c}{12}\\
Training steps & 1M & 500K & 3M \\
Batch size & 256 & 5112& 256 \\
Learning rate & 1e-4 & 4e-4 & 1e-4 \\
Warmup steps & 10,000 & 24,000& 4,000 \\
Adam $\beta$ & (0.9, 0.99) & (0.9, 0.98)& (0.9, 0.99) \\
Adam $\epsilon$ & \multicolumn{3}{c}{1e-6} \\
Learning rate schedule & \multicolumn{3}{c}{Linear} \\
Weight decay & \multicolumn{3}{c}{0.01} \\
Dropout & \multicolumn{3}{c}{0.1} \\
GPUs (Nvidia V100) & 16 & 64& 32 \\
\bottomrule
\end{tabular}
}
\caption{Hyperparameters used for pre-training on different text corpora.}
\label{table:pt-hparam}
\end{table}


\section{Hyperparameters for Fine-Tuning}
The hyperparameters for each tasks are searched on the development sets according to the average score of ten runs with different random seeds.

\subsection{GLUE benchmark}
The fine-tuning hyper-parameters  for GLUE benchmark~\cite{glue} are presented in Table~\ref{table:f_glue}. 
\begin{table}[H]

\centering
\setlength{\belowcaptionskip}{-0.3cm}
\small
\resizebox{0.48\textwidth}{!}{
\begin{tabular}{lr}
\toprule
\textbf{Hyperparameters} & \textbf{GLUE}  \\ \midrule
Batch size & \{16, 32\}  \\
Learning rate & \{5e-5, 1e-4, 1.5e-4\}  \\
Epochs & 3 for MNLI and \{10, 15\} for others \\
LR schedule & Linear  \\
Layerwise LR decay & 0.8\\
Warmup proportion & 0.1 \\
Weight decay & 0.01  \\
\bottomrule
\end{tabular}
}
\caption{Hyperparameters used for fine-tuning on the GLUE benchmark.}
\label{table:f_glue}
\end{table}

\subsection{SQuAD benchmark and RACE dataset}
The fine-tuning hyper-parameters for SQuAD~(\citealp{squad1};\citealp{squad2}) and RACE~\cite{race} are presented in Table~\ref{table:f_squad}. 
\begin{table}[H]
\setlength{\belowcaptionskip}{-0.2cm}
\centering
\small
\renewcommand\tabcolsep{2.5pt}
\begin{tabular}{lrr}
\toprule
\textbf{Hyperparameters} & \textbf{SQuAD}  & \textbf{RACE}\\ 
\midrule
Batch size & 48&32  \\
Learning rate & \{1e-4, 1.5e-4, 2e-4\}& \{8e-5, 1e-4\}\\
Epochs &\{2, 4\}&\{4, 5\}\\
LR schedule & Linear & Linear \\
Layerwise LR decay & 0.8& 0.8\\
Warmup proportion & 0.1& 0.1 \\
Weight decay & 0.0& 0.01 \\
\bottomrule
\end{tabular}
\caption{Hyperparameters used for fine-tuning on the SQuAD benchmark and RACE dataset.}
\label{table:f_squad}
\end{table}

\subsection{Text Classification tasks}
Table~\ref{table:f_imdb} lists the fine-tuning hyper-parameters for IMDb~\cite{imdb} and AG'news~\cite{ag} datasets. To process texts with a length larger than $512$, we follow~\citealp{fine-tune_long} to select the first $512$ tokens to perform fine-tuning.
\begin{table}[H]
\centering
\setlength{\belowcaptionskip}{-0.2cm}
\small
\renewcommand\tabcolsep{2.pt}
\begin{tabular}{lrr}
\toprule
\textbf{Hyperparameters} & \textbf{IMDb} &\textbf{AG'news}\\ 
\midrule
Batch size & \multicolumn{2}{c}{32}  \\
Learning rate & \multicolumn{2}{c}{\{5e-5, 1e-4, 1.5e-4\}}  \\
Epochs & \multicolumn{2}{c}{3} \\
LR schedule & \multicolumn{2}{c}{Linear}  \\
Layerwise LR decay & \multicolumn{2}{c}{0.8}\\
Warmup proportion & \multicolumn{2}{c}{0.1} \\
Weight decay & \multicolumn{2}{c}{0.01}  \\
\bottomrule
\end{tabular}
\caption{Hyperparameters used for fine-tuning on IMDb and AG'news.}
\label{table:f_imdb}
\end{table}

\subsection{Chinese NLU tasks}
The fine-tuning hyperparameters for Chinese NLU tasks including XNLI~\cite{xnli}, LCQMC~\cite{lcqmc}, DRCD~\cite{drcd}, DuReader~\cite{dureader}, CMRC2018 and MSRA-NER~\cite{ner} are presented in Table~\ref{table:f_chinese}. 
\begin{table}[H]
\centering
\setlength{\belowcaptionskip}{-0.3cm}
\small
\renewcommand\tabcolsep{2.pt}
\begin{tabular}{lcccc}
\toprule
\multirow{2}{*}{\textbf{Tasks}}  &\textbf{Batch}&\textbf{Learning}&\multirow{2}{*}{\textbf{Epoch}}&\multirow{2}{*}{\textbf{Droput}}\\
&\textbf{size}&\textbf{rate}&&\\ 
\midrule
XNLI &256&1.5e-4&3&0.1  \\
LCQMC &32&4e-5&2&0.1  \\
CMRC2018&64&1.5e-4&5&0.2 \\
DuReader&64&1.5e-4&5&0.1  \\
DRCD&64&1.5e-4&3&0.1\\
MSRA-NER&16&1.5e-4&10&0.1 \\
\bottomrule
\end{tabular}
\caption{Hyperparameters used for fine-tuning on Chinese NLU tasks. Note that all tasks use the layerwise lr decay with decay rate $0.8$.}
\label{table:f_chinese}
\end{table}

\begin{table*}[htb]
\centering
\renewcommand\tabcolsep{2.8pt}
\resizebox{0.98\textwidth}{!}{
\begin{tabular}{l|c|cccccccc|c|cccc}
\toprule
\multirow{2}{*}{\textbf{Models}} & \textbf{Size of} & \textbf{MNLI} & \textbf{QNLI} & \textbf{QQP} & \textbf{SST-2} & \textbf{CoLA} & \textbf{MRPC} & \textbf{RTE} & \textbf{STS-B} & \textbf{GLUE}&\multicolumn{2}{c}{\textbf{SQuAD1.1}} &\multicolumn{2}{c}{\textbf{SQuAD2.0}}\\ 
&\textbf{Lexicon}&Acc&Acc&Acc&Acc&MCC&Acc&Acc&PCC&\textbf{Avg}&EM&F1&EM&F1 \\
\midrule
BERT$_{\rm Reimplement}$&$0$K&84.9&91.8&91.3&92.9&58.8&88.1&69.7&88.6&83.4&83.4&90.2&76.4&79.2 \\
\midrule
&$100$K&85.4&92.3&91.3&92.9&60.4&88.7&72.6&89.6&84.1&84.2&90.8&78.4&81.5 \\ 
Contiguously&$200$K&85.3&92.0&91.5&92.7&59.3&89.0&71.5&89.5&83.9&84.2&90.9&78.3&81.3 \\
MLM&$300$K&85.1&92.1&91.3&92.8&59.3&88.6&73.3&89.5&84.0&83.9&90.7&78.5&81.4 \\
&$400$K&85.0&92.0&91.3&93.1&58.3&89.2&71.8&89.1&83.7&83.9&90.7&78.0&81.1 \\
\midrule
&$100$K&85.3&92.2&91.4&92.9&62.3&88.6&72.5&88.0&84.2&84.2&90.9&78.6&81.4 \\
Explicitly&$200$K&85.4&92.3&91.3&92.8&62.1&88.4&74.5&88.6&84.4&84.5&\textbf{91.3}&78.9&81.9 \\
N-gram MLM&$300$K&85.7&92.3&91.3&92.9&62.6&88.7&75.8&89.4&\textbf{84.8}&\textbf{84.7}&91.2&\textbf{79.5}&\textbf{82.4} \\
&$400$K&85.3&92.2&91.4&92.9&61.3&88.5&73.2&89.3&84.3&84.6&\textbf{91.3}&79.0&81.7	\\
\bottomrule
\end{tabular}
}
\caption{Results on the development set of the GLUE and SQuAD benchmarks with different MLM objectives and diverse sizes of $n$-gram lexicon.}
\label{table:appendix_lexicon_size}
\end{table*}

\begin{table*}[htb]
\centering
\renewcommand\tabcolsep{2.8pt}
\resizebox{1.\textwidth}{!}{
\begin{tabular}{cl|ccccccccc|c}
\toprule
\multirow{2}{*}{$\texttt{\#}$}&\multirow{2}{*}{\textbf{Models}}  & \multicolumn{2}{c}{\textbf{MNLI}} & \textbf{QNLI} & \textbf{QQP} & \textbf{SST-2} & \textbf{CoLA} & \textbf{MRPC} & \textbf{RTE} & \textbf{STS-B} & \textbf{GLUE}\\ 
&&m&mm&Acc&Acc&Acc&MCC&Acc&Acc&PCC&\textbf{Avg} \\
\midrule
$\texttt{\#1}$&ERNIE-Gram$_{{\rm BASE}}$&87.1&87.1&92.8&91.8&93.2&68.5&90.3&79.4&90.4&\textbf{86.7} \\ 
$\texttt{\#2}$&$\texttt{\#1}-$\ relative~position~bias&86.5&86.4&92.5&91.6&93.2&68.1&90.3&79.4&90.6&86.5\\
$\texttt{\#3}$&$\texttt{\#2}-$\ comprehensive~$n$-gram~prediction ({\tt CNP})&86.2&86.2&92.4&91.7&92.7&65.5&90.0&78.7&90.5&86.0 \\
$\texttt{\#4}$&$\texttt{\#2}-$\ enhanced~$n$-gram~relation~modeling ({\tt ENRM})&85.7&85.8&92.6&91.2&93.5&64.8&88.9&76.9&90.0&85.5 \\
$\texttt{\#5}$&$\texttt{\#4}-$\ comprehensive~$n$-gram~prediction ({\tt CNP})&85.6&85.7&92.3&91.3&92.9&62.6&88.7&75.8&89.4&84.8 \\
\bottomrule
\end{tabular}
}
\caption{Comparisons between several ERNIE-Gram variants on GLUE benchmark. All the listed models are pre-trained following the same settings of BERT$_{{\rm BASE}}$~\cite{bert}.}
\label{table:append_ablation}
\end{table*}

\begin{table*}[h]
\small
\centering
\renewcommand\tabcolsep{5pt}
\begin{tabular}{l|ccccccc}
\toprule
\multirow{2}{*}{\textbf{Models}}  & \multicolumn{2}{c}{\textbf{MNLI}}  & \textbf{SST-2} & \multicolumn{2}{c}{\textbf{SQuAD1.1}} & \multicolumn{2}{c}{\textbf{SQuAD2.0}}  \\ 
&m&mm&Acc&EM&F1&EM&F1 \\
\midrule
MPNet~\cite{mpnet}&86.2&-&\textbf{94.0}&85.0&91.4&80.5&83.3 \\
$-{\rm relative~position~bias}$&85.6&-&93.6&84.0&90.3&79.5&82.2\\
\textsc{UniLM}v2~\cite{unilm}&86.1&86.1&93.2&85.6&92.0&80.9&83.6\\
$-{\rm relative~position~bias}$&85.6&85.5&93.0&85.0&91.5&78.9&81.8\\
\midrule
ERNIE-Gram&\textbf{87.1}&\textbf{87.1}&93.2&\textbf{86.2}&\textbf{92.3}&\textbf{82.1}&\textbf{84.8}\\
$-{\rm relative~position~bias}$&86.5&86.4&93.2&85.2&91.7&80.8&84.0\\
\bottomrule
\end{tabular}

\caption{Ablation study on relative position bias~\cite{t5} for ERNIE-Gram and previous strong pre-trained models like MPNet and \textsc{UniLM}v2.}
\label{table:ablation_pos}
\end{table*}
\begin{figure*}[!h]
\setlength{\belowcaptionskip}{-0.5cm}
\begin{center} 
\includegraphics[width=0.8\linewidth]{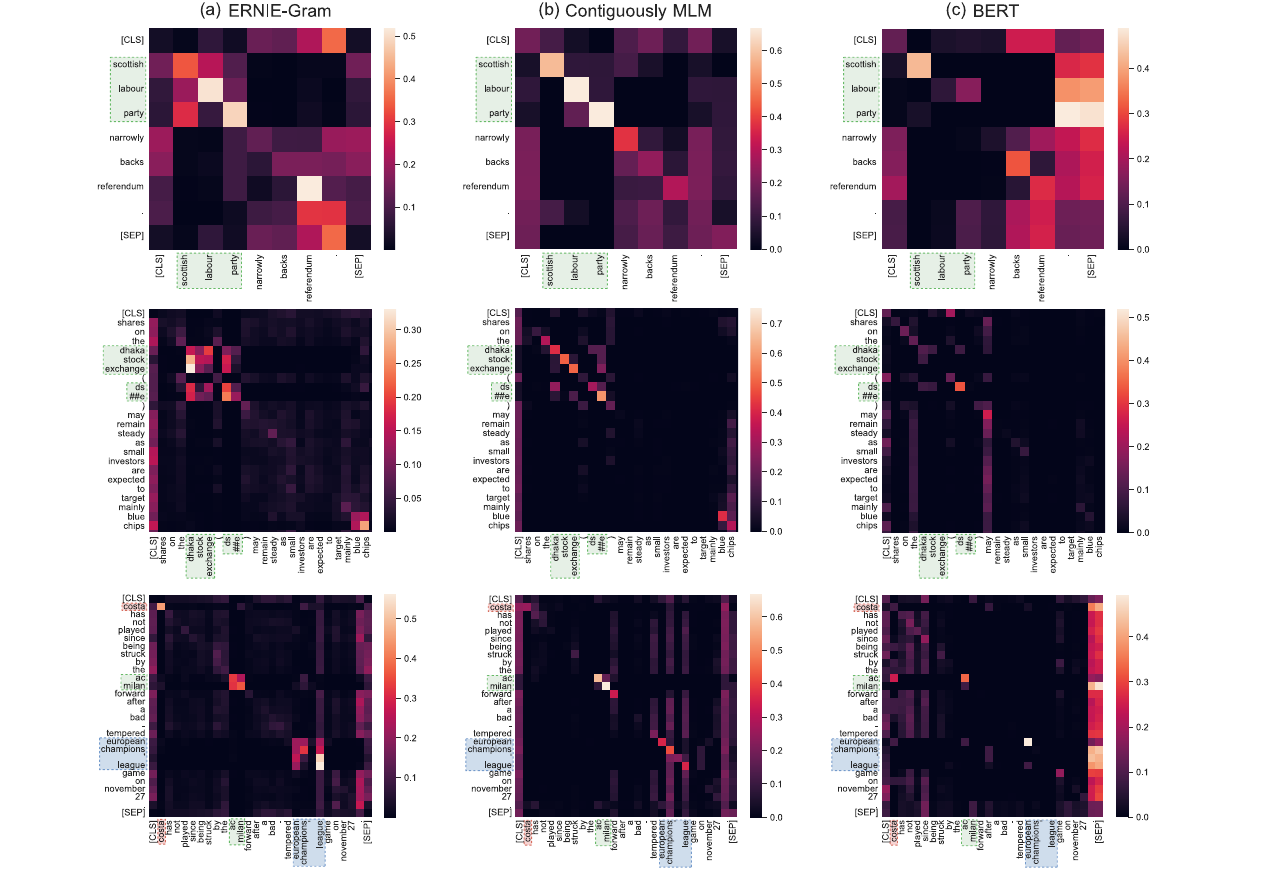}
\caption{(a-c) Mean attention scores in the last self-attention layer. Texts in \colorbox{mygreen1}{\makebox(18,6){\strut green}}, \colorbox{myorange1}{\makebox(24,6){\strut orange}}, \colorbox{myred}{\makebox(12,6){\strut red}} and \colorbox{myblue1}{\makebox(16,6){\strut blue}} boxes are named entities standing for organizations, locations, person and miscellaneous respectively.}
\label{cases}
\end{center} 
\end{figure*}

\section{Detailed Results for Ablation Studies}
\label{appedix_c}
We present the detailed results on GLUE benchmark for ablation studies in this section. The results on different MLM objectives and sizes of $n$-gram lexicon are presented in Table~\ref{table:appendix_lexicon_size}. The detailed results on ERNIE-Gram variants to verify the effectiveness of comprehensive $n$-gram prediction and enhanced $n$-gram relation modeling mechanisms are presented in Table~\ref{table:append_ablation}. 
Results of ablation study on relative position bias~\cite{t5} are presented in Table~\ref{table:ablation_pos}.

\section{More cases on CoNLL2003 Dataset}
\label{appedix_d}
We visualize the attention patterns of three supplementary cases from CoNLL2003 named entity recognition dataset~\cite{conll03} to compare the performance of ERNIE-Gram, contiguously MLM and BERT (lowercased), as shown in Figure~\ref{cases}. For contiguously MLM, there are clear diagonal lines in named entities that tokens prefer to attend to themselves. While for ERNIE-Gram, there are bright blocks over named entities that tokens attend to most of tokens in the same entity adequately to construct tight representation.

\end{document}

%% file: settings.tex
\usepackage{multirow}
\usepackage{amsmath}
\usepackage{capt-of}
\usepackage{tabularx}
\usepackage[caption=false]{subfig}
\usepackage{epsfig}
\usepackage{amssymb}
\usepackage{amsfonts}
\usepackage{booktabs}
\usepackage{scalerel}
\usepackage[inline]{enumitem}
\usepackage{listings}
\usepackage{varwidth}
\usepackage[export]{adjustbox}
\usepackage{tikz}
\usetikzlibrary{tikzmark}
\usepackage{cleveref}

\usepackage{stmaryrd}
\usepackage{bbm}

\usepackage{algorithm}
\usepackage[noend]{algpseudocode}

\definecolor{deepblue}{rgb}{0,0,0.5}
\definecolor{officeblue}{RGB}{0,102,204}
\definecolor{deepred}{rgb}{0.6,0,0}
\definecolor{deepgreen}{rgb}{0,0.5,0}
\definecolor{mybrickred}{RGB}{182,50,28}

\definecolor{fillcolor}{RGB}{216,217,252}

\algnewcommand\algorithmicrequireb{{\hspace{0.85cm}}}
\algnewcommand\INPTDESCB{\item[\algorithmicrequireb]}

\algnewcommand\algorithmicfuncdesc{\textbf{Function:}}
\algnewcommand\FUNCDESC{\item[\algorithmicfuncdesc]}
\algnewcommand\algorithmicfuncdescb{{\hspace{1.48cm}}}
\algnewcommand\FUNCDESCB{\item[\algorithmicfuncdescb]}
\algnewcommand{\algorithmicgoto}{\textbf{goto}}
\algnewcommand{\Goto}[1]{\algorithmicgoto~\ref{#1}}

\newcommand*\AlgCommentInLine[1]{{\color{deepblue}{$\triangleright$ \textit{#1}}}}


%% file: math_commands.tex
\usepackage{amsmath,amsfonts,bm}

\def\eqref#1{equation~\ref{#1}}

\def\1{\bm{1}}

\DeclareMathAlphabet{\mathsfit}{\encodingdefault}{\sfdefault}{m}{sl}
\SetMathAlphabet{\mathsfit}{bold}{\encodingdefault}{\sfdefault}{bx}{n}

